\documentclass{article} 
\usepackage{iclr2024_conference,times}

\usepackage{soul}

\usepackage{pdfpages}
\usepackage{amsmath}
\usepackage{amsfonts}
\usepackage{mathtools}
\usepackage{amsthm}

\usepackage{microtype}
\usepackage{graphicx}
\usepackage{booktabs} 
\usepackage{algorithm}
\usepackage{algorithmic}
\usepackage[backref=page,colorlinks=true,linkcolor=blue,citecolor=green,urlcolor=blue]{hyperref}
\usepackage{url}
\usepackage{graphicx}
\usepackage{bbm}
\usepackage{placeins}
\usepackage{adjustbox}
\usepackage{xcolor}
\usepackage{scalerel}
\usepackage{natbib}
\usepackage[outdir=./]{epstopdf}
\usepackage{graphicx,float,pgfplots,wrapfig,sidecap,lipsum}

\usepackage{colortbl}
\usepackage{diagbox} 
\usepackage{tablefootnote}
\usepackage[font=small,labelfont=bf]{caption}
\usepackage{subcaption}
\usepackage{subfloat}
\usepackage{tikz}
\usetikzlibrary{fit}
\usetikzlibrary{calc,shapes}
\usetikzlibrary{decorations.pathmorphing} 
\usetikzlibrary{fit}					
\usetikzlibrary{backgrounds}	
\usetikzlibrary{pgfplots.groupplots}

\usepackage[utf8]{inputenc}
\usepackage{pgfplots}
\usepgfplotslibrary{groupplots,dateplot}
\usetikzlibrary{patterns,shapes.arrows}
\pgfplotsset{compat=newest}

\usepackage{xspace}
\usepackage{soul}
\usepackage{booktabs}
\usepackage{bm}
\usepackage{listings}

\theoremstyle{remark}

\AtBeginEnvironment{proof}{}{}{}

\definecolor{sourcecolor}{rgb}{0.5,1,0.5}
\definecolor{ourcolor}{rgb}{1,0,0}
\definecolor{singlecolor}{rgb}{0,0,1}
\definecolor{auxcolor}{rgb}{0.54,0.17,0.89}
\definecolor{linearcolor}{rgb}{0.172549019607843,0.627450980392157,0.172549019607843}
\definecolor{randomcolor}{rgb}{1,0.498039215686275,0.0549019607843137}
\definecolor{tunecolor}{rgb}{0.9568627450980393, 0.8156862745098039,0}
\definecolor{aligncolor}{rgb}{0,0.5,0}

\DeclareMathOperator*{\argmax}{arg\,max}

\title{Robustness to Multi-Modal Environment Uncertainty in MARL using Curriculum Learning}

\author{Aakriti Agrawal$^{1}$, Rohith Aralikatti$^{1}$, Yanchao Sun$^{1}$ \& Furong Huang$^{1}$ \\
$^{1}$Department of Computer Science, University of Maryland \\
\texttt{\{agrawal5,rohithca,ycs,furongh\}@umd.edu} \\
}
\iclrfinalcopy 
\begin{document}

\maketitle
\begin{abstract}
Multi-agent reinforcement learning (MARL) plays a pivotal role in tackling real-world challenges. However, the seamless transition of trained policies from simulations to real-world requires it to be robust to various environmental uncertainties. Existing works focus on finding Nash Equilibrium or the optimal policy under uncertainty in one environment variable (i.e. action, state or reward). This is because a multi-agent system itself is highly complex and unstationary. However, in real-world situation uncertainty can occur in multiple environment variables simultaneously. This work is the first to formulate the generalised problem of robustness to multi-modal environment uncertainty in MARL. To this end, we propose a general robust training approach for multi-modal uncertainty based on curriculum learning techniques. We handle two distinct environmental uncertainty simultaneously and present extensive results across both cooperative and competitive MARL environments, demonstrating that our approach achieves state-of-the-art levels of robustness.
\end{abstract}

\section{Introduction}

Multi-Agent Reinforcement Learning (MARL) has shown tremendous success in solving many complex real-world decision making problems such as in robotics(path-planning \citep{path-planning}, task-allocation \citep{agrawal2023rtaw, agrawal2022dc}), traffic management \citep{traffic}; Game Theory and Economics \citep{nowe2012game} etc. In MARL \citep{marl_rev}, the agent interacts with the environment as well the other agents to maximize its own long-term return in a shared environment. Thus making it more complex than single agent RL with finding Nash Equilibrium as one of most popular solution concept \citep{marl_ne}.

Real-world MARL applications require training the agent in simulations and transferring it to the real-world. In such a situation, its possible that the agent does not have accurate knowledge or there is shift in the environment parameters (eg. reward function/transition dynamics function is not exactly same as in the simulations) or there is noise (eg. the information about state, action, reward are not transferred with perfect accuracy) or issues in the hardware, etc. Thus, leading to environment uncertainty. Though there has been some work on robustness to uncertainty in reward, transition dynamics \citep{model}, state \citep{state_per} and action \citep{action} but they have been studied individually. Real world setting will not foretell the exact uncertain parameter and will have uncertainty in multiple environment parameters thus requiring the need to develop robustness to multi-modal uncertainty. 



\textbf{Contributions.} In this work, \textbf{(1)} we focus on developing robustness to two uncertain parameter at a time, thus, making this the first work to handle multi-modal uncertainty in MARL. \textbf{(2)} We also define and theoretically formalise the general problem of finding optimal policy in MARL with multi-modal uncertainty. \textbf{(3)} Generally, finding optimal policy in MARL is characterized by finding Nash Equilibrium for the markov game however fining NE for this generalised problem is highly complex. In this work, we formally investigate and design an efficient curriculum learning (CL) to solve this problem. \textbf{(4)} We also show experimentally that our method is able to find optimal solution for the given robust Markov games and generate state-of-the-art robustness for reward, state and action uncertainty. \textbf{(5)} As a by product, this is the also the first work to handle action uncertainty in MARL.

\begin{figure}[!htbp]
    \centering    \includegraphics[width=\linewidth]{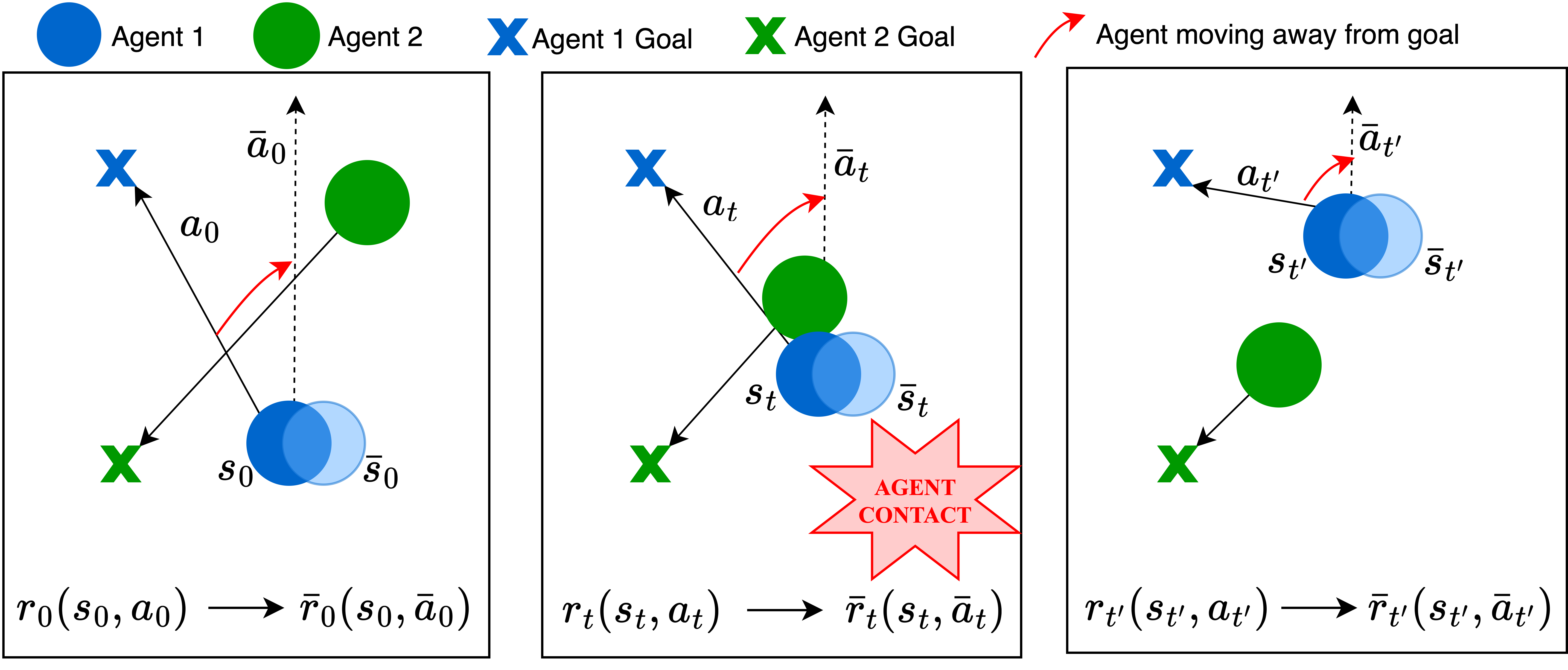}
    \caption{In this figure we try to motivate the problem of solving multi-modal uncertainty for MARL. We show here some of the problems that can occur due to state, action and reward uncertainty. For example, at time = 0 (first image) agent 2 observes the perturbed state $\bar{s}_{0}$ of agent 1 and not the true state $s_{0}$, leading to a collision at the time = t (middle image). Agent 1 needs to reach the goal but its perturbed actions take it away from the goal (last image). Reward uncertainty does not give accurate expected return for an action in a given state, thus making the training more difficult.}
    \label{fig:motivation}
\end{figure}

\section{Related Work}
\vspace{-2mm}
\textbf{Robustness in RL.} Robustness in reinforcement learning is due to adversarial attacks \citep{adv} or uncertainty in model/environment parameters. In single agent RL, robustness to uncertainty is handled by maximin optimisation between the agent and the uncertainty set in the form of zero sum game \citep{RL_action, RL_robust_mdp, RL_robust2, robust_mdp3}. In MARL, uncertainty is defined in the form of robust Markov game. The agents individually maximise their return while interacting with each other in presence of uncertainty. MARL research has made tremendous advancements in the recent times, however there is only very few work handling uncertainty. \cite{marl_uncern} is the only work on multi-modal uncertainty (reward and transition dynamics) but it only handles it theoretically. Some of the robustness to single uncertainty works handle reward, transition dynamics~\citep{model} or state \citep{state_per} uncertainty. \cite{m3ddpg, romax} develop robustness to the adversarial agent's actions by training the agent to handle the worst case action of the adversarial agents in a competitive setting. Robustness to action perturbations in cooperative environment with adversaries had been studied in \cite{romaq, robust_coop, robust_coop2}.

\textbf{Curriculum learning} \cite{CL, CLRLSurvey} has been widely used for improving robustness in various different domains, such as object classification \citep{obj_class}, automatic speech recognition \citep{asr}, etc. \textbf{CL for robustness in RL.} \cite{CL_RL} is the closest work to ours on using CL to improve robustness in RL. They implement bootstrapped opportunistic curriculum learning to improve robustness in single agent RL with RADIAL-DQN as baseline. \cite{CL_generalization} develops robustness by developing generalisation to multiple tasks. They formulate the training curriculum as a multi-armed bandit problem which selects the task to train the RL model to give maximum reward gain. 

There is no existing work which uses CL for multi-modal uncertainty. It is also first time it is being used to handle robustness in MARL.


\section{Problem Formulation}
This section gives the problem formulation with multiple uncertainties (reward, state, action and transition dynamics) and also introduce the different uncertainties in RL. 
\subsection{Robust markov game}
The interaction among multiple agents can be modeled as \textbf{markov game} $\mathcal{G}$ \citep{markov_game}, which can be defined as a tuple, 
$$ \mathcal{G} =  \langle \mathcal{N}, \{ \mathcal{S}^{i} \}_{i \in \mathcal{N}}, \{ \mathcal{A}^{i} \}_{i \in \mathcal{N}}, \{ \mathcal{R}^{i}_{s} \}_{(i,s) \in \mathcal{N} \times \mathcal{S}}, \{ \mathcal{P}_{s} \}_{s \in \mathcal{S}}, \gamma \rangle .$$

Here $\mathcal{N} = [N]$ denotes the set of N agents, $\mathcal{S}^{i}$ and $\mathcal{A}^{i}$ denotes the state space and action space of agent $i \in \mathcal{N}$ respectively. 
$\mathcal{S} = \mathcal{S}^{1} \times \dots \times \mathcal{S}^{N}$ is the joint state space. 
$\mathcal{R}^{i} \colon \mathcal{S} \times \mathcal{A}^1 \times \dots\times \mathcal{A}^{N} \rightarrow \mathbb{R}$ represents the reward function of agent $i$, which depends on the current state and the joint action of all agents. 
$\mathcal{P} :S \times A^{1} \times \dots \times A^{N} \rightarrow \Delta(S)$ represents the state transition probability that is a mapping from the current state and the joint action to the probability distribution over the state space. 
$\gamma \in [0, 1)$ is the discounting factor. At time t, agent $i$ chooses its action $a^{i}_{t}$ according to policy $\pi^{i}: \mathcal{S}^{i} \rightarrow \Delta(A^{i})$. 
The agents' joint policy $\pi = \Pi_{i \in \mathcal{N}} \pi^{i} : \mathcal{S} \rightarrow \Delta (\mathcal{A})$.

In real-world situation a markov game will not have accurate information available and thus contain uncertainties in model such as reward, state, action and/or transition probability model. Thus we define \textbf{Robust markov game} \citep{RewardTransition2011}  as, 
\begin{equation}
    \mathcal{\bar{G}}_{general} =  \langle \mathcal{N}, \{ \mathcal{S}^{i} \}_{i \in \mathcal{N}}, \{ \mathcal{\bar{O}}^{i} \}_{i \in \mathcal{N}}, \{ \mathcal{A}^{i} \}_{i \in \mathcal{N}},  \{ \bar{\mathrm{A}}^{i} \}_{i \in \mathcal{N}},  \{ \mathcal{\bar{R}}^{i}_{s} \}_{(i,s) \in \mathcal{N} \times \mathcal{S}}, \{ \mathcal{\bar{P}}_{s} \}_{s \in \mathcal{S}},  \gamma \rangle,
    \label{eq:robust_markov}
\end{equation}
where $\mathcal{N}$, $\mathcal{S}^{i}$, $\mathcal{A}^{i}$, and $\gamma \in [0,1)$ denote the set of agents, the state space, the action space for each agent $i$, and the discounting factor, respectively. $\mathcal{\bar{R}}^{i}_{s} \in \mathbb{R}^{|A|}$ and $\mathcal{\bar{P}}_{s}$ denotes the uncertainty sets of all possible reward function values and that of all possible transition probabilities at state $s$. $\mathcal{\bar{O}}^{i}$ and $\bar{\mathrm{A}}^{i}$ denotes the uncertainty sets of all possible values of perturbed state $\bar{s}^{i}$ and $\bar{a}^{i}$ respectively. The state space and action space of the uncertainty sets is similar to that of the true state $s$ and true action $a$ respectively. 

Note that the state perturbation reflects the state uncertainty from the perspective of each agent, but it does not change the true state of multi-agent systems. The state still transits from the true state to the next state. Each agent is associated with a policy $\pi^{i} : S \rightarrow \Delta(A^{i})$ to choose an action $a^{i} \in A^{i}$ given the perturbed state $\bar{s}$.




\subsection{Uncertainty in RL}
In this section we explain the different types of uncertainty in RL and introduce our uncertainty model for each. 

Uncertainty can be categorised into two kinds, aleatoric and epistemic. Aleatoric uncertainty or statistical uncertainty originates from the statistic nature of the environment and interactions with the environment. This uncertainty can be modeled and evaluated but cannot be reduced. Whereas epistemic uncertainty or model uncertainty originates from current limitations of training the neural network and is reducible. There are 3 main potential sources of aleatoric uncertainty in RL \citep{uncern} which corresponds to the three main components of the MDP, stochastic rewards, stochastic observations, and stochastic actions. However, stochastic observations can occur due to stochastic transition dynamics as well as stochastic observations itself. 

\textbf{Stochastic rewards} would mean that for every state and action pair, we now have a distribution of the reward rather than a fixed reward. Rewards only play a role during training and not during testing. Thus developing a robust system to reward uncertainties imply that we are able to train the model to converge and achieve the goal even with high reward uncertainty. The model we have used for \textbf{stochastic reward function} is defined as: 
\begin{equation}
    \bar{R}^{i}(s,a) = \mathcal{N}_{trunc}(R^{i}(s,a), \epsilon),
    \label{eq:reward}
\end{equation}
where $R^{i}$ is true reward for agent $i$, $\bar{R}^{i}$ is perturbed reward, $\epsilon$ is standard deviation and $\mathcal{N}_{trunc}$ is truncated normal distribution truncated at 2$\epsilon$. Increasing $\epsilon$ will increase the uncertainty. 

The \textbf{stochastic observations} can stem from stochastic transition dynamics or stochastic observations itself. If the $\mathcal{P}$ function in the MDP is non-deterministic, then the transition from one state to the next is a source of uncertainty. Thus, \textbf{stochastic transition dynamics} means that for every current state and action pair, we now have a distribution over the next state and not a fixed specific next state. 

In the other scenario of stochastic observations, the true state of the system remains unchanged and only the observed state is perturbed. The input to the policy network is the perturbed state, but for the reward and transition dynamics functions input is the true state. The model we have used for \textbf{stochastic observation function} is: 
\begin{equation}
    \bar{s}^{i} = \mathcal{N}_{trunc}(s^{i}, \mu),
\label{eq:state}
\end{equation}
where $s$ is true state, $\bar{s}$ the perturbed state, $\mu$ is standard deviation and $\mathcal{N}_{trunc}$ is truncated normal distribution truncated at 2$\mu$. Increasing $\mu$ will increase the uncertainty. 

The \textbf{stochastic actions} means that there is uncertainty in about the next state due to uncertain actions. One example is any stochastic policy algorithm (PPO, SAC) in which the action is chosen from a distribution instead of a deterministic point. We define our model for stochastic actions as: 
\begin{equation}
    \bar{a}^{i} = \mathcal{N}_{trunc}(a^{i}, \nu),
\label{eq:action}
\end{equation}
where $a$ is true action, $\bar{a}$ the perturbed action, $\nu$ is standard deviation and $\mathcal{N}_{trunc}$ is truncated normal distribution truncated at 2$\nu$. Increasing $\nu$ will increase the uncertainty.

Note that the range of value of observations and actions is quite small as compared to that of reward. Thus, the magnitude of robustness is different for different uncertainty parameters.

\section{Robust Nash Equilibrium with multimodal uncertainty} 

\subsection{Nash Equilibrium in MARL}
NE is one of the commonly used solution concepts in multi-agent static games. We will now introduce NE in MARL
The expected return in case of multi agent RL for $i^{th}$ agent is
$$V^{i}_{\pi}(s) = \mathbb{E} [ {\sum_{t=0}^{\infty}} \gamma^{t} r^{i}_{t} | s_{0} = s, a^{i}_{t} \sim \pi^{i}(.|s_{t}), a^{-i}_{t} \sim \pi^{-i}(.|s_{t}) ],$$
where $-i$ represents the indices of all agents except agent $i$, and $\pi^{-i} = \Pi_{i \neq j} \pi_{j}$ refers to the joint policy of all agents except agent $i$. In order to find the optimal robust value function for the single agent the other agent policies are considered stationary. Since all policies are evolving continuously and expected return is dependent on all agent policies, one commonly used solution for optimal policy $\pi^{*} = \{\pi_{1}^{*}, \pi_{2}^{*}, \dots \pi_{N}^{*} \}$ is nash equilibrium. $\pi_{*}$ is called Markov perfect Nash Equilibrium. Optimal value function is defined by,

\begin{equation}
    V^{i}_{*}(s) = \max_{\pi^{i}(.|s)} \sum_{a \in \mathcal{A}} \prod_{j=1}^{N} \pi^{j}(a^{j}|s^{j})(R^{i}(s, a) + \gamma \sum_{s'\in S} P(s'|s, a) V^{i}_{*}(s')).
    \label{eq:bellan_marl}
\end{equation}

Non-stationarity is one of the main reasons for difficulty in MARL convergence, which is further attenuated with uncertainty.

\subsection{Robust Nash Equilibrium in MARL with multi-modal uncertainty}
In this section, we define the solution for the general robust Markov game in equation \ref{eq:robust_markov} which includes the 4 kinds of aleatoric uncertainty in a MARL system. Uncertainty in one model parameter directly or indirectly affects the others, but there could be additional stochasticity in other parameters in the real world. Ideally, we would like to make our model robust to all four model uncertainties, but handling uncertainty in all 4 variables is a complex problem. Therefore, most existing work focuses only on single uncertainty (state \cite{state_per}, action\cite{action}, reward \cite{model}). This is majorly due to the complexity of finding Nash equilibrium and optimal Bellman equation.

We now define the Bellman equation for the value function including all four uncertainties discussed in the previous section - state, action, reward, and transition dynamics. We follow the maximin approach of optimization where we minimize the bellman equation \ref{eq:bellan_marl} for each agent $i$  with respect to the four uncertainty sets and maximize with respect to its policy $\pi^{i}$. Our aim is to select the entries $\bar{P}$, $\bar{R}^{i}_{s}$, $\bar{s}$, $\bar{a}$, from uncertainty set $\mathcal{\bar{P}}_{s}$, $\mathcal{\bar{R}}^{i}_{s}$, $\mathcal{\bar{O}}$, $\mathrm{\bar{A}}$ that minimises the expected return. Thus, the optimal Bellman equation will be, 
$$\bar{V}_{*}^{i}(s^{i}) = \max_{\pi^{i}(.|s^{i})} \min_{\begin{array}{c}\scriptstyle \bar{P}(.|s,.) \in \mathcal{\bar{P}}_{s} \\[-4pt] \scriptstyle \bar{R}^{i}_{s} \in \mathcal{\bar{R}}^{i}_{s} \\[-4pt] \scriptstyle \bar{s} \in \mathcal{\bar{O}} \\[-4pt] \scriptstyle \bar{a} \in \mathrm{\bar{A}} \end{array}} \sum_{a \in \mathcal{A}} \prod_{j=1}^{N} \pi^{j}(a^{j}|\bar{s}^{j})(\bar{R}^{i}(s, \bar{a}) + \gamma \sum_{s'\in S} \bar{P}(s'|s, \bar{a}) \bar{V}_{*}^{i}(s')),$$
where $\bar{s} = \{ \bar{s}^{1}, \bar{s}^{2}, ... \bar{s}^{N} \}$ and  $\bar{a} = \{ \bar{a}^{1}, \bar{a}^{2}, ... \bar{a}^{N} \}$. The true state of an agent remains unchanged but only the state perceived by other agents is perturbed, therefore only policy $\pi^{i}$ takes perturbed state $\bar{s}^{i}$ as input whereas reward $R$ and transition dynamics $P$ function takes true state $s$ as input. The policy $\pi^{i}(.|\bar{s}^{i})$ then produces the true actions $a$ which are then perturbed by the environment into $\bar{a}$. The reward $R$ and transition dynamics $P$ function takes perturbed $\bar{a}$ as input and eventually gets perturbed itself. 

If an optimal value function exists, then we define the existence of robust Nash equilibrium (RNE). RNE is the solution for the robust Markov game.

\textbf{Definition 1:} (Robust Nash Equilibrium) Given a Markov game $\mathcal{\bar{G}}_{general}$ with state, reward, action and transition dynamics uncertainty, a joint policy $\pi_{*} = \{ \pi_{*}^{1}, \pi_{*}^{2} \dots \pi_{*}^{N} \}$ is said to be robust Nash equilibrium for $ i \in \mathcal{N}$, $s \in \mathcal{S}$, iff there exists optimal value function $V_{*} = \{ V_{*}^{1}, V_{*}^{2}, \dots V_{*}^{N}\}$ and satisfies, 

$$\pi^{i}_{*}(.|s^{i}) \in \argmax_{\pi^{i}(.|s^{i})} \min_{\begin{array}{c}\scriptstyle \bar{P}(.|s,.) \in \mathcal{\bar{P}}_{s} \\[-4pt] \scriptstyle \bar{R}^{i}_{s} \in \mathcal{\bar{R}}^{i}_{s} \\[-4pt] \scriptstyle \bar{s} \in \mathcal{\bar{O}} \\[-4pt] \scriptstyle \bar{a} \in \mathrm{\bar{A}} \end{array}} \sum_{a \in \mathcal{A}} \pi^{i}(a^{i}|\bar{s}^{i}) \prod_{i \neq j} \pi^{j}_{*}(a^{j}|\bar{s}^{j})(\bar{R}^{i}(s, \bar{a}) + \gamma \sum_{s'\in S} \bar{P}(s'|s, \bar{a}) \bar{V}_{*}^{i}(s')).$$

\textbf{Theorem 1:} Existence of robust Nash Equilibrium $\rightarrow$ Existence of optimal Value Function.

For detailed proof check appendix \ref{theorem1}. This proof has been conducted for reward and transition dynamics uncertainty \cite{model} but not for partially observable games. It has also been explored for state uncertainty \cite{state_per}. We focus on developing the general proof when all possible uncertainties are present in MARL. 

Theoretically proving the existence of NE policy for the $\mathcal{\bar{G}}_{general}$ is out of the scope of this work. \cite{state_per} find NE for state uncertainty, also shown in the appendix. \ref{NE_state} and \cite{model} find NE for reward/transition dynamics uncertainty, also shown in appendix \ref{NE_reward}.

\section{Curriculum Learning based robustness to multi-modal uncertainty}
\textbf{Curriculum learning} \citep{CL, CLRLSurvey} is a methodology to optimize the order in which experience is accumulated by the agent, so as to increase performance or training speed on a set of final tasks. 
An important challenge while designing the curriculum for training is developing a strategy to measure the difficulty level of our task. In our case, we use the noise parameter ($\epsilon$, $\mu$ and $\nu$) to increase the task difficulty level. Through generalization, knowledge acquired by training in simple tasks can be leveraged to reduce the exploration of more complex tasks. Therefore by training for lower levels of noise parameter, the model learns faster for higher levels of noise. Thus making our method sample efficient. 

Our base model for applying CL is \cite{model} and we introduce state, reward, and action uncertainty in it as shown in equations \ref{eq:state}, \ref{eq:reward}, \ref{eq:action}  respectively.

\subsection{Efficient Lookahead CL}
This section presents our efficient curriculum learning algorithm for the case when only one uncertain parameter (state, reward, or action) is present. There are a total of four possible uncertain parameters, however, most work considers uncertainty in transition dynamics and state to be similar and does not study it separately. We do not have transition dynamics uncertainty because our transition dynamics are deterministic.  The algorithm is described in \ref{alg1}.

\begin{algorithm}
\caption{Lookahead CL for single uncertainty}
\label{alg1}
\begin{algorithmic}
\REQUIRE $\Delta \lambda$ where $\lambda \in \{\epsilon, \mu, \nu \}$
\STATE $\lambda \leftarrow 0$
\STATE $\lambda_{converged} \leftarrow False$
\STATE $TrainTillSuccess(\lambda)$
\WHILE{\NOT $\lambda_{converged}$}
\STATE $\lambda \leftarrow \lambda + \Delta \lambda$
\STATE $\lambda_{converged} \leftarrow TrainTillSuccess(\lambda)$
\ENDWHILE
\end{algorithmic}
\end{algorithm}

\subsection{Efficient Lookahead CL for multiple uncertainties}
The algorithm for efficient curriculum learning for reward with state/action perturbations is described in \ref{alg:algo1}. 
Our aim is to simultaneously increase both reward and state/action uncertainty in an efficient manner in each iteration of curriculum learning, to eventually train a model that can handle two uncertainties at the same time. Note there is no skip ahead in the case of the reward uncertainty parameter since we do not have reward uncertainty during the evaluation phase. In a similar way algorithm for action and reward perturbations is shown in algorithm \ref{alg:algo2}.

\begin{minipage}{.5\textwidth}
%
  \begin{algorithm}[H]
    \caption{
    Reward+State/Action Uncertainty}
    \label{alg:algo1}
    \begin{algorithmic}
    \REQUIRE $\Delta \epsilon, \Delta \mu/\nu$
    \STATE $\epsilon \leftarrow 0$, $\mu/\nu \leftarrow 0$
    \STATE $\epsilon_{converge} \leftarrow False$, \\ $\mu_{converge} \leftarrow False$
    \STATE $TrainToSucc(\epsilon, \mu/\nu)$
    \STATE $\mu \leftarrow SkipAhead(\mu/\nu)$
    \WHILE{\NOT $(\epsilon_{converge} \; \AND \; \mu/\nu_{converge})$}
    \IF{\NOT $\epsilon_{converge}$}
    \STATE $\epsilon \leftarrow \epsilon + \Delta \epsilon$
    \STATE $\epsilon_{converge} \leftarrow TrainToSucc(\epsilon, \mu/\nu)$
    \STATE 
    \ENDIF
    \IF{\NOT $\mu_{converge}$}
    \STATE $\mu \leftarrow \mu + \Delta \mu/\nu$
    \STATE $\mu_{converge} \leftarrow TrainToSucc(\epsilon, \mu/\nu)$
    \STATE $\mu \leftarrow SkipAhead(\mu/\nu)$
    \ENDIF
    \ENDWHILE
    \end{algorithmic}
\end{algorithm}
\end{minipage}%
\begin{minipage}{.5\textwidth}
  \begin{algorithm}[H]
    \caption{
    State+Action Uncertainty}
    \label{alg:algo2}
    \begin{algorithmic}
    \REQUIRE $\Delta \nu, \Delta \mu$
    \STATE $\nu \leftarrow 0$, $\mu \leftarrow 0$
    \STATE $\nu_{converge} \leftarrow False$, \\ $\mu_{converge} \leftarrow False$
    \STATE $TrainToSucc(\nu, \mu)$
    \STATE $\mu \leftarrow SkipAhead(\mu)$
    \WHILE{\NOT $(\nu_{converge} \; \AND \; \mu_{converge})$}
    \IF{\NOT $\nu_{converge}$}
    \STATE $\nu \leftarrow \nu + \Delta \nu$
    \STATE $\nu_{converge} \leftarrow TrainToSucc(\nu, \mu)$
    \STATE $\nu \leftarrow SkipAhead(\nu)$ 
    \\
    \ENDIF
    \IF{\NOT $\mu_{converge}$}
    \STATE $\mu \leftarrow \mu + \Delta \mu$
    \STATE $\mu_{converge} \leftarrow TrainToSucc(\nu, \mu)$
    \STATE $\mu \leftarrow SkipAhead(\mu)$
    \ENDIF
    \ENDWHILE
    \end{algorithmic}
    \end{algorithm}
\end{minipage}

\section{Experiments}
In this section, we show results for our curriculum learning-based method on three multi-particle environments and compare the results with state-of-the-art robustness in those environments. The three environments are cooperative navigation, keep-away, and physical deception. For each of these environments, we first compare results for the base method (without CL) with our CL method for varying levels of either reward, state, or action uncertainty. Then we show results for the combination of two uncertainties: state + reward, action + reward, and state + action. This paper is the first to handle multi-modal uncertainty in MARL. We note values related to reward and state uncertainty by evaluating the trained model 1000 times and reporting its mean and standard deviation. However, for reward uncertainty, we show training plots since rewards do not play a role in evaluation and therefore uncertainty during evaluation does not make sense. We show

\vspace{-1mm}
\subsection{Robustness under uncertainty in a single parameter.}

\begin{figure}[H]
    \centering
    \begin{subfigure}{0.3\linewidth}
        \includegraphics[width=\linewidth]{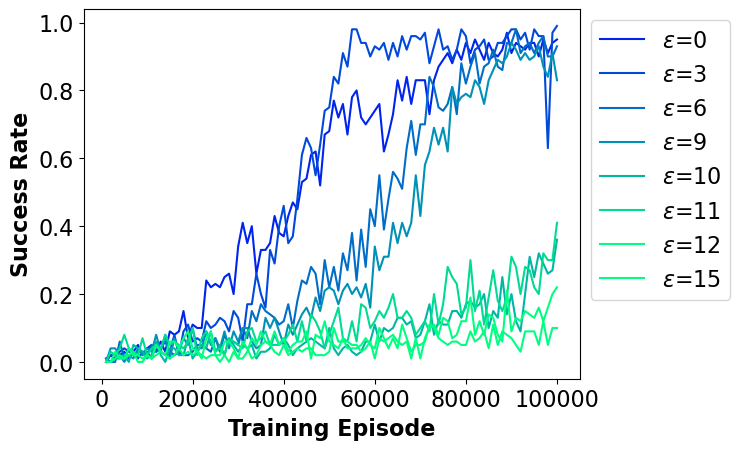}
        \caption{Success rate for various $\epsilon$ in reward uncertainty.}
        \label{fig:sr_reward_simple}
    \end{subfigure}
    \hfill
    \begin{subfigure}{0.3\linewidth}
        \includegraphics[width=\linewidth]{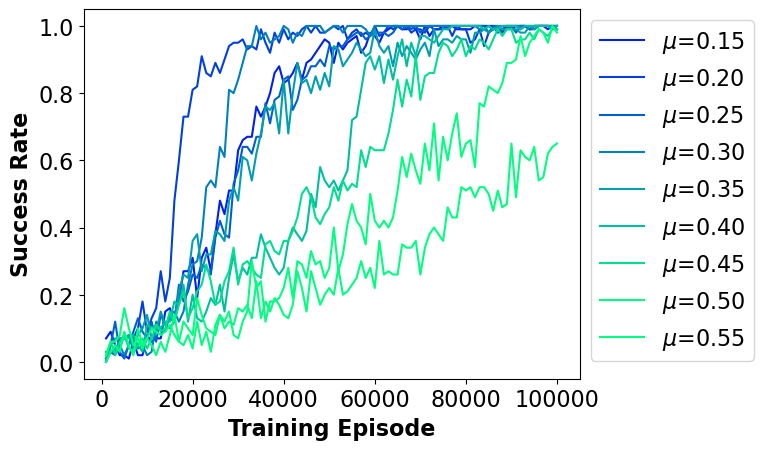}
        \caption{Success rate for various $\mu$ in state uncertainty.}
        \label{fig:sr_state_simple}
    \end{subfigure}
    \hfill
    \begin{subfigure}{0.3\linewidth}
        \includegraphics[width=\linewidth]{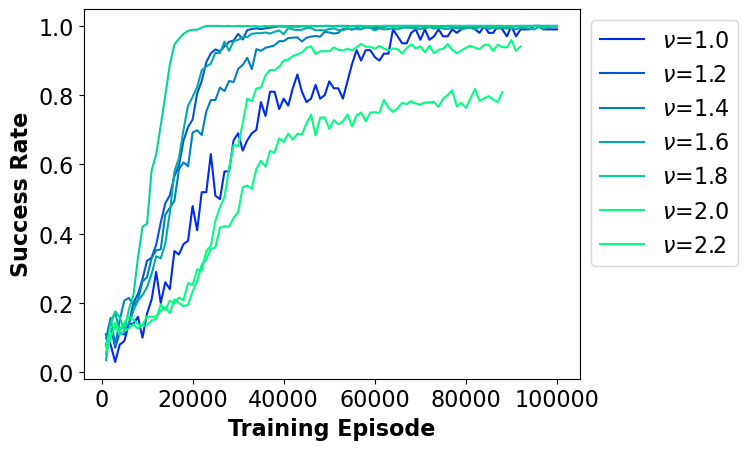}
        \caption{Success rate for various $\nu$ in action uncertainty.}
        \label{fig:sr_action_simple}
    \end{subfigure}
    \caption{\textbf{Cooperative Navigation: Baseline Performance.} Success rate vs training time. An episode is successful if all landmarks are occupied by all agents. Reward uncertainty shows good performance until $\epsilon$=9 (left), state uncertainty shows good performance until $\mu$=0.5 (middle) and action uncertainty shows good performance until $\nu$=2.0 (right).}
\end{figure}

\begin{figure}[!htbp]
    \centering
    \begin{subfigure}{0.3\linewidth}
        \includegraphics[width=\linewidth]{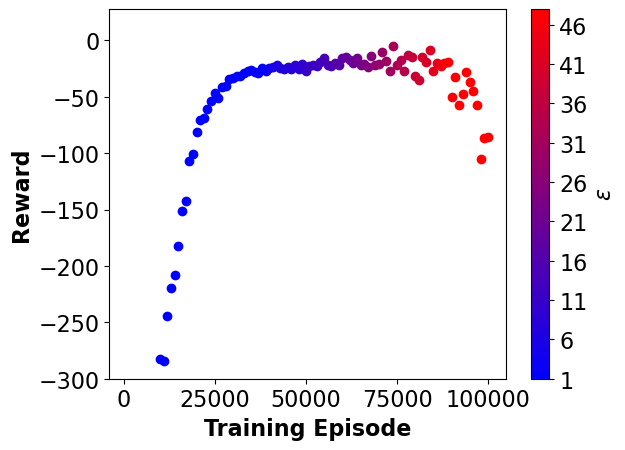}
        \caption{Reward Uncertainty.}
        \label{fig:rew_reward}
    \end{subfigure}
    \hfill
    \begin{subfigure}{0.3\linewidth}
        \includegraphics[width=\linewidth]{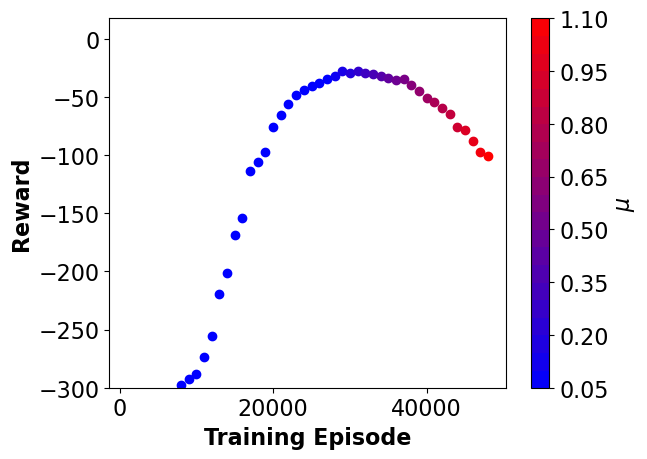}
        \caption{State Uncertainty.}
        \label{fig:rew_state}
    \end{subfigure}
    \hfill
    \begin{subfigure}{0.3\linewidth}
        \includegraphics[width=\linewidth]{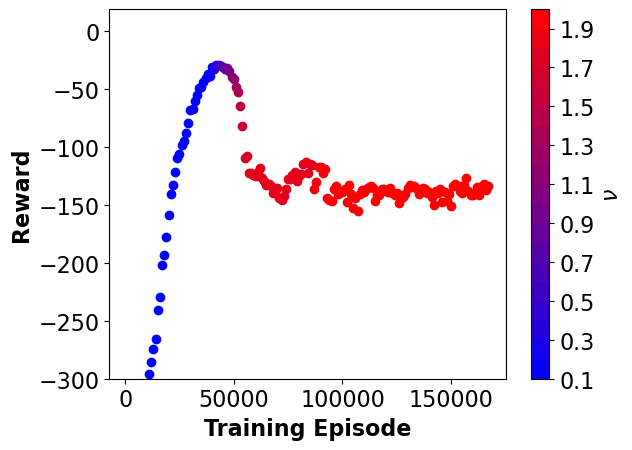}
        \caption{Action Uncertainty.}
        \label{fig:rew_action}
    \end{subfigure}
    \caption{\textbf{Cooperative Navigation: CL Method Performance.} This plot shows the changing reward as the noise value is incremented in the CL method for the three uncertain parameters separately. Reward uncertainty learns until $\epsilon$=47 (left), state uncertainty until $\mu$=1.1 (middle), and action uncertainty learns until $\nu$=2.2 (right).}
\end{figure}

\begin{table}[H]
\resizebox{\columnwidth}{!}{%
\begin{tabular}{|c|c c c c c c c c|}
\hline
\multicolumn{9}{|c|}{Reward Uncertainty} \\
\hline
$\epsilon$ & 6       & 9         & 10     & 11    & 12       & 15           & 47        & 48    \\
Baseline  &  $\checkmark$ & $\checkmark$ & $\times$ & $\times$ & $\times$ &    $\times$  & $\times$ & $\times$ \\
CL  & $\checkmark$    & $\checkmark$       & $\checkmark$ & $\checkmark$ & $\checkmark$    & $\checkmark$ & $\checkmark$ & $\times$ \\
\hline
\multicolumn{9}{|c|}{State Uncertainty} \\
\hline
$\mu$ & 0.25 & 0.45 & 0.5 & 0.55 & 0.75 & 1.0 & 1.1 & 1.2 \\
Baseline  & 1 $\pm$ 0.04 & 0.97 $\pm$ 0.15   & \textbf{1 $\pm$ 0.1} & 0.6 $\pm$ 0.5 & - &  - & - &  - \\
CL  & 1 $\pm$ 0  & 1 $\pm$ 0 & 1 $\pm$ 0  & 1 $\pm$ 0 & 1 $\pm$ 0.1 & 0.98 $\pm$ 0.2 & \textbf{0.94 $\pm$ 0.2} & 0.82 $\pm$ 0.4 \\

\hline
\multicolumn{9}{|c|}{Action Uncertainty} \\
\hline
$\nu$    & 1.0           & 1.2     & 1.4     & 1.6        & 1.8           & 2.0           & 2.2   & 2.4 \\
Baseline   & 1 $\pm$ 0.13 & 1 $\pm$ 0 & 1 $\pm$ 0 & 1 $\pm$ 0 & 1 $\pm$ 0.1     & 0.93 $\pm$ 0.26 & 0.78 $\pm$ 0.41 & - \\
CL     & 1 $\pm$ 0       & 1 $\pm$ 0 & 1 $\pm$ 0 & 1 $\pm$ 0    & 1 $\pm$ 0.1 & 1 $\pm$ 0 & \textbf{1 $\pm$ 0.03} & 0.9 $\pm$ 0.04 \\

\hline
\end{tabular}%
}
\caption{\textbf{Cooperative Navigation.} The table shows a detailed comparison between CL and baseline for reward uncertainty (top), state uncertainty (middle), and action uncertainty (bottom) using success rates. For reward uncertainty, the CL-based method learned to converge until $\epsilon$ = 48, whereas baseline learned until $\epsilon$ = 9. For state uncertainty, the CL-based method gave a high success rate until $\mu$ = 1.1, whereas baseline learned until $\mu$ = 0.5. For action uncertainty, the CL-based method gave a high success rate until $\nu$ = 2.4, whereas baseline learned until $\nu$ = 2.0.}
\label{tab:Coop_nav_action_state}
\end{table}

\textbf{Cooperative Navigation Environment.} This is a cooperative environment where three agents learn to occupy all three landmarks while avoiding collisions. For \textbf{reward uncertainty} we see from the success rate figure \ref{fig:sr_reward_simple} that the model can learn until $\epsilon = 9$ which is the current state-of-the-art robustness \citep{model}. However, with our lookahead-CL method, we were able to achieve robustness up to $\epsilon = 47$ (check reward plot \ref{fig:rew_reward}). \textbf{For state uncertainty} as we see from the figure \ref{fig:sr_state_simple} the model was able to learn until $\mu = 0.5$ while with CL it reached $\mu = 1.1$ (check reward plot \ref{fig:rew_state}). \textbf{For action uncertainty} we see from the figure \ref{fig:sr_action_simple} the model was able to learn until $\nu = 2.0$ while with CL it reached $\nu = 2.2$ (check reward plot \ref{fig:rew_action}). This is the first work on action uncertainty so we do not have a baseline to compare with. For state uncertainty, \cite{state_per} does not have a comparison for various uncertainty values. Table \ref{tab:Coop_nav_action_state} shows detailed comparison results between CL and baseline, and eventually concludes that CL achieves state-of-the-art robustness.

\subsection{Multi-modal CL combining two uncertainty:}
\begin{figure}[htp]
    \centering
    \begin{subfigure}{0.45\linewidth}
        \includegraphics[width=\linewidth]{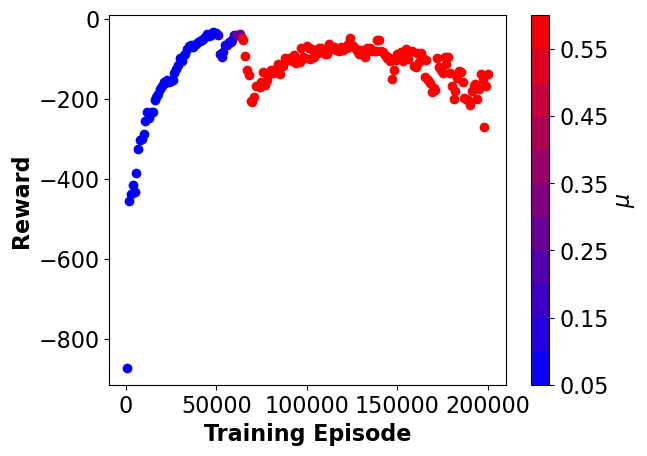}
        \caption{Reward changes with $\mu$}
    \end{subfigure}
    \hfill
    \begin{subfigure}{0.45\linewidth}
        \includegraphics[width=\linewidth]{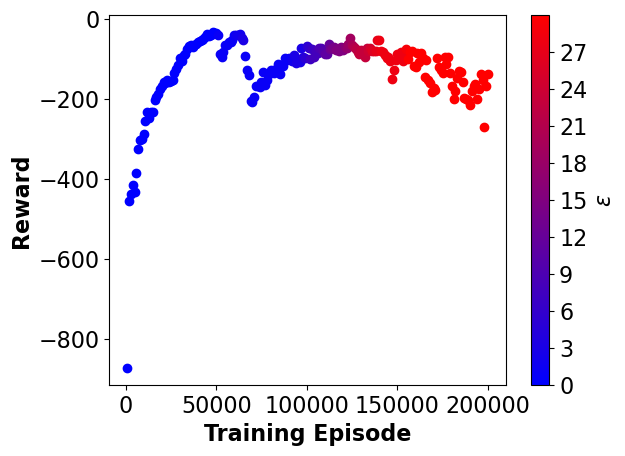}
        \caption{Reward changes with $\epsilon$}
    \end{subfigure}
    
    \caption{Reward plot for lookahead CL for the case of multiple uncertainties (reward and state) showing the changing reward for various $\mu$ (left) and $\epsilon$ (right).}
    \label{fig:rewstate}
\end{figure}

\begin{figure}[htp]
    \centering
    \begin{subfigure}{0.45\linewidth}
        \includegraphics[width=\linewidth]{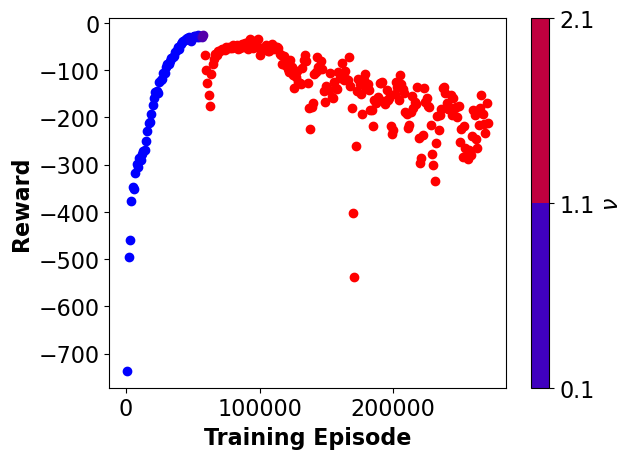}
        \caption{Reward changes with $\nu$}
    \end{subfigure}
    \hfill
    \begin{subfigure}{0.45\linewidth}
        \includegraphics[width=\linewidth]{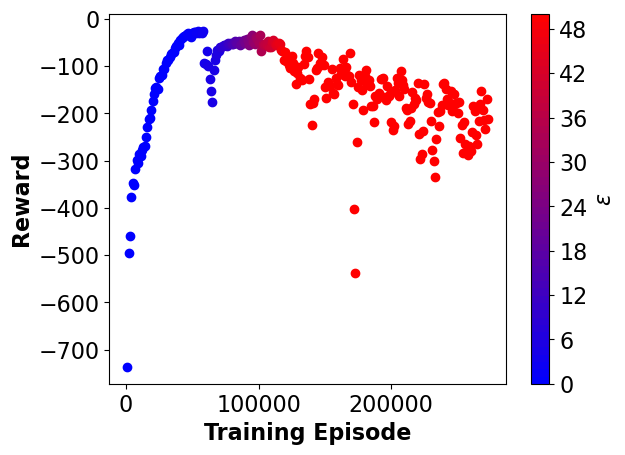}
        \caption{Reward changes with $\epsilon$}
    \end{subfigure}
    
    \caption{Reward plot for lookahead CL for the case of multiple uncertainties (reward and action) showing the changing reward for various $\nu$ (left) and $\epsilon$ (right).}
    \label{fig:rewaction}
\end{figure}

\begin{figure}[htp]
    \centering
    \begin{subfigure}{0.45\linewidth}
        \includegraphics[width=\linewidth]{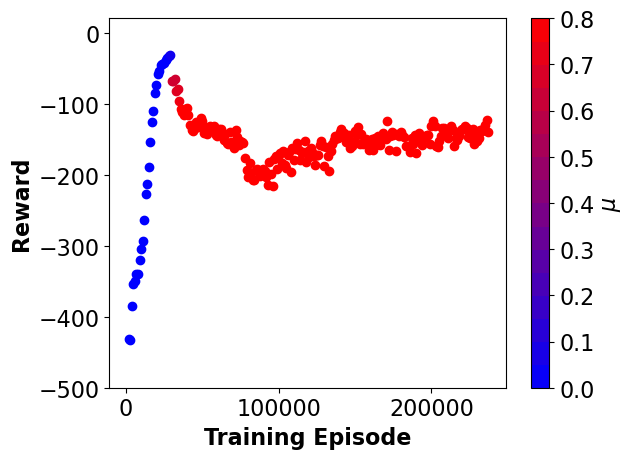}
        \caption{Reward changes with $\mu$}
    \end{subfigure}
    \hfill
    \begin{subfigure}{0.45\linewidth}
        \includegraphics[width=\linewidth]{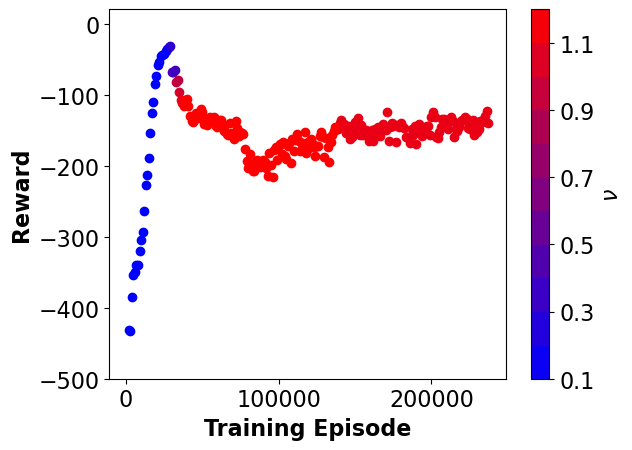}
        \caption{Reward changes with $\nu$}
    \end{subfigure}
    
    \caption{Reward plot for lookahead CL for the case of multiple uncertainties (state and action) showing the changing reward for various $\mu$ (left) and $\nu$ (right).}
    \label{fig:actionstate}
\end{figure}

\textbf{Cooperative Navigation Environment:} We try the combinations,  reward + state,  reward + action, and action + state uncertainty to test how CL works when two uncertainties are combined. 
For \textbf{reward+state} uncertainty combination as shown in table \ref{tab:coop_nav_CL} we are able to learn until $\mu$ = 0.7 when $\epsilon$ = 0 and $\epsilon$ = 29 while training Check figure \ref{fig:rewstate} for reward plots. For \textbf{reward+action} uncertainty combination as shown in table \ref{tab:coop_nav_CL} we are able to learn uptil $\nu$ = 2.4 when $\epsilon$ = 0 and $\epsilon$ = 50 while training. Check figure \ref{fig:rewaction} for reward plots. For \textbf{action+state} uncertainty combination as shown in table \ref{tab:coop_nav_CL} we are able to learn uptil $\nu$ = 3 when $\mu$ = 0 and $\mu$ = 1 when $\nu$ = 0. Check figure \ref{fig:actionstate} for reward plots. Thus, we can see that even after introducing two uncertainties at the same time, the method outperforms the baseline. 

\begin{table}[htp]
\resizebox{\columnwidth}{!}{%
\begin{tabular}{|lllllllll|}
\hline
\multicolumn{9}{|c|}{Reward + State Uncertainty}                                                                   \\ \hline
\multicolumn{1}{|l|}{$\mu$}  & 0.5        & 0.55          & 0.6           & 0.65 & \textbf{0.7} & 0.75 & 0.8 & 0.85 \\
\hline
\multicolumn{1}{|l|}{} & 0.97 $\pm$ 0.2 & 0.95 $\pm$ 0.2   &  0.95 $\pm$ 0.2  & 0.96 $\pm$ 0.2 & \textbf{0.95 $\pm$ 0.2} & 0.94 $\pm$ 0.23 & 0.91 $\pm$ 0.3 &  0.81 $\pm$ 0.4 \\ \hline
\multicolumn{9}{|c|}{Reward + Action Uncertainty}                                                                \\ \hline
\multicolumn{1}{|l|}{$\nu$} & 1 & 1.2  & 1.4 & 1.6  & 1.8 & 2.0  & 2.2 & \textbf{2.4}    \\
\multicolumn{1}{|l|}{}             &    0.96 $\pm$ 0.2   & 0.96 $\pm$ 0.2   & 0.97 $\pm$ 0.2     & 0.97 $\pm$ 0.2 & 0.97 $\pm$ 0.2 & 0.96 $\pm$ 0.2 & 0.96 $\pm$ 0.2 & \textbf{0.96 $\pm$ 0.2}    \\ \hline
\multicolumn{9}{|c|}{State+Action Uncertainty} \\
\hline

\multicolumn{1}{|l|}{$\mu$ - $\nu$} & 0             & 1    & 1.4 & 1.6           & 2             & 2.4         & 2.6 & 3    \\
\hline
\multicolumn{1}{|l|}{0}             &  1 $\pm$ 0       & 1 $\pm$ 0.6 & 1 $\pm$ 0.04 & 1 $\pm$ 0.03    & 1 $\pm$ 0.04   & \textbf{1 $\pm$ 0.05}  & \textbf{1 $\pm$ 0.06}   & \textbf{0.96 $\pm$ 0.18} \\
\multicolumn{1}{|l|}{0.6}             &  1 $\pm$ 0.6 & 0.98 $\pm$ 0.1 & 0.98 $\pm$ 0.14 & \textbf{0.96 $\pm$ 0.2} & \textbf{0.92 $\pm$ 0.28} & 0.8 $\pm$ 0.4 & 0.73 $\pm$ 0.4 & 0.6 $\pm$ 0.5 \\
\multicolumn{1}{|l|}{0.8}             &    0.98 $\pm$ 0.14 & \textbf{0.95 $\pm$ 0.2} & \textbf{0.9 $\pm$ 0.3}   & 0.85 $\pm$ 0.35 & 0.8 $\pm$ 0.45  & 0.6 $\pm$ 0.5 & 0.5 $\pm$ 0.5  & 0.4 $\pm$ 0.5
\\
\multicolumn{1}{|l|}{1}             &  \textbf{0.93 $\pm$ 0.25} & 0.8 $\pm$ 0.4  & 0.75 $\pm$ 0.43 & 0.67 $\pm$ 0.5  & 0.6 $\pm$ 0.5   &  0.42 $\pm$ 0.5          & 0.36 $\pm$ 0.5 & 0.26 $\pm$ 0.23
\\

\hline
\end{tabular}%
}
\caption{\textbf{Cooperative Navigation.} This table shows the success rate performance of the CL-based method when both reward and state uncertainty (top), reward and action uncertainty (middle), or state and action uncertainty (bottom) are present. An episode is successful if all landmarks are occupied by all agents. }
\label{tab:coop_nav_CL}
\end{table}

\begin{figure}[!htbp]
    \centering
    \begin{subfigure}{0.3\linewidth}
        \includegraphics[width=1\linewidth]{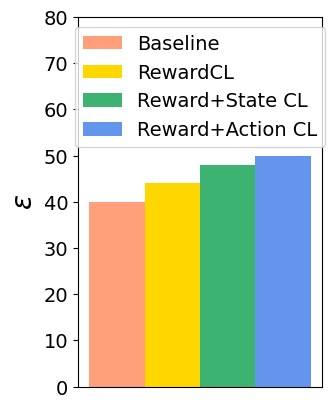}
        \caption{Highest $\epsilon$ for which convergence is achieved for different training strategies.}
        \label{fig:reward_barplot}
    \end{subfigure}
    \hfill
    \centering
    \begin{subfigure}{0.3\linewidth}
        \includegraphics[width=1\linewidth]{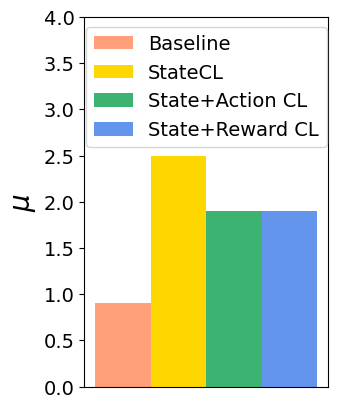}
        \caption{Highest $\mu$ for which convergence is achieved for different training strategies.}
        \label{fig:state_barplot}
    \end{subfigure}
    \hfill
    \centering
    \begin{subfigure}{0.3\linewidth}
        \includegraphics[width=0.95\linewidth]{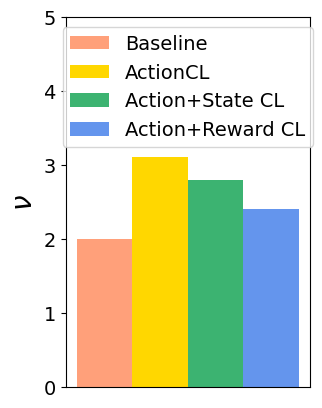}
        \caption{Highest $\nu$ for which convergence is achieved for different training strategies.}
        \label{fig:action_barplot}
    \end{subfigure}
    \caption{\textbf{Keep Away.} The highest noise values that lead to convergence (agent success rate is greater than 90\%) are shown above. We see that CL applied to a single parameter does better than the baseline training with no CL. When CL is applied to multiple parameters, the highest noise value is usually lower than applying CL to a single parameter. However, the same model now supports uncertainty in two different parameters for a slight reduction in the noise value. }
    \label{fig:barplots_keep_away}
\end{figure}

\textbf{Keep Away environment:} This is  a competitive environment with one agent, one adversary, and two landmarks. One landmark is randomly chosen to be the goal location and the objective of the agent is to reach the goal and that of the adversary is to block the agent from reaching the goal. The agent knows which landmark is the goal while the adversary does not. For all experiments in this environment, noise is added only to the agent's state, reward, and action. Evaluation is also done similarly. An agent is considered successful if it reaches the goal within 100 steps.

\begin{figure}[H]
    \centering
    \begin{subfigure}{0.45\linewidth}
        \includegraphics[width=\linewidth]{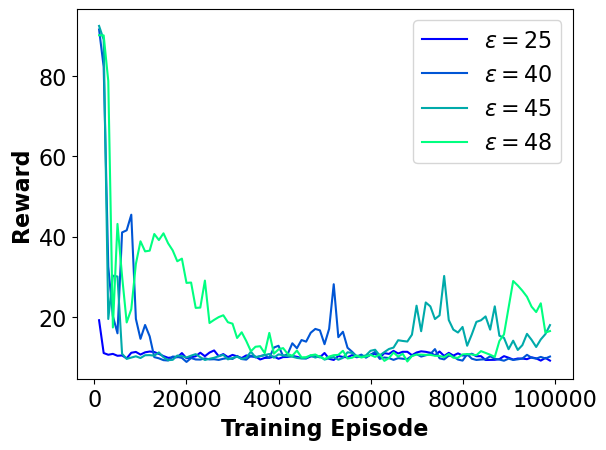}
        \caption{Time taken by agent to reach the goal for various $\epsilon$ in reward uncertainty.}
    \end{subfigure}
    \hfill
    \begin{subfigure}{0.45\linewidth}
        \includegraphics[width=\linewidth]{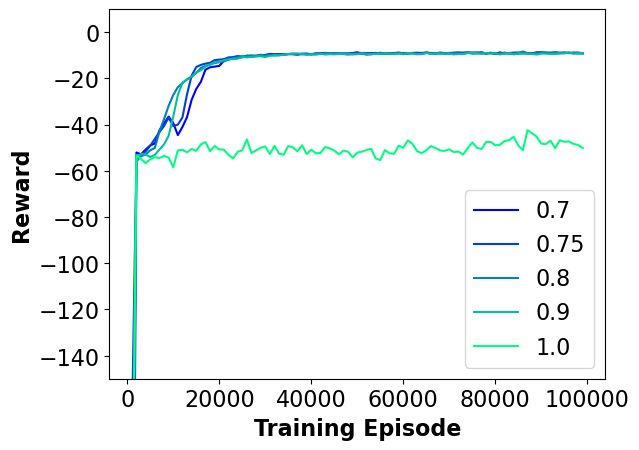}
        \caption{Reward changes for various $\mu$ in state uncertainty.}
    \end{subfigure}
    \hfill
    \begin{subfigure}{0.45\linewidth}
        \includegraphics[width=\linewidth]{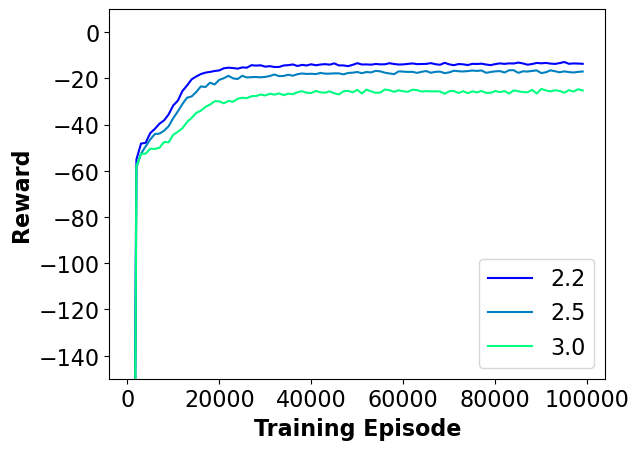}
        \caption{Reward changes for various $\nu$ in action uncertainty.}
    \end{subfigure}
    \caption{\textbf{Keep Away: Baseline Performance.} For reward uncertainty we show the plot between number of steps taken by an agent to reach the goal vs training time. This is because due to reward uncertainty reward is noisy and hence a plot of noisy reward will not give good conclusions. We observe that this number saturates for $\epsilon$ = 40 but for number higher that this, its heavily fluctuating hence concluding that reward uncertainty learns until $\epsilon$ = 40. For state and action uncertainty we show reward vs training time. State uncertainty shows good performance until $\mu$=0.9 (middle) and action uncertainty shows good performance until $\nu$=2.0 (last).}
    \label{fig:ka_baseline}
\end{figure}

\begin{figure}[H]
    \centering
    \begin{subfigure}{0.45\linewidth}
        \includegraphics[width=\linewidth]{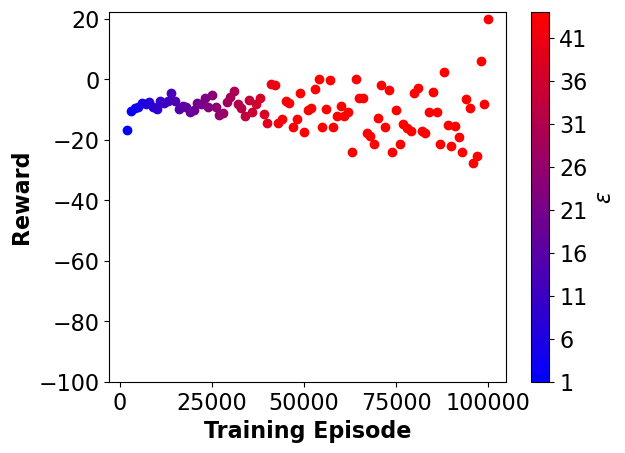}
        \caption{Reward Uncertainty.}
    \end{subfigure}
    \hfill
    \begin{subfigure}{0.45\linewidth}
        \includegraphics[width=\linewidth]{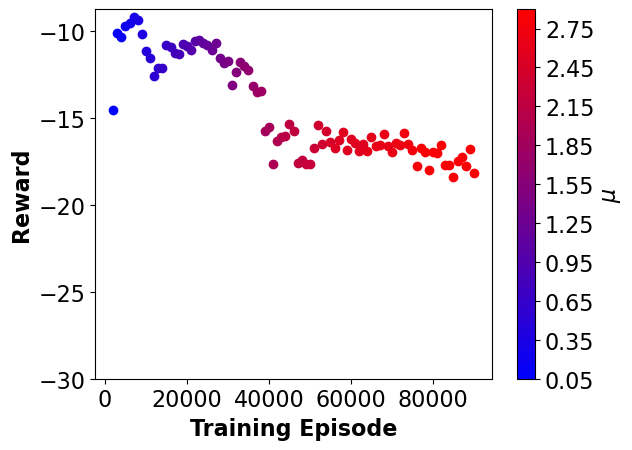}
        \caption{State Uncertainty.}
    \end{subfigure}
    \hfill
    \begin{subfigure}{0.45\linewidth}
        \includegraphics[width=\linewidth]{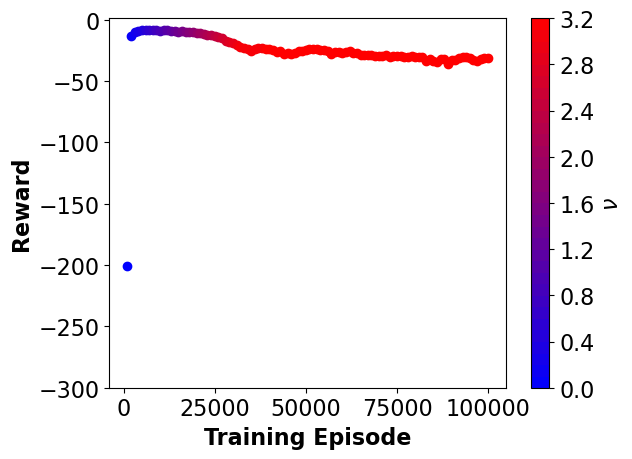}
        \caption{Action Uncertainty.}
    \end{subfigure}
    \caption{\textbf{Keep Away: CL Method Performance.} This plot shows the changing reward as the noise value is incremented in the CL method for the three uncertain parameters separately. Reward uncertainty learns until $\epsilon$=43 (left), state uncertainty until $\mu$=2.5 (middle), and action uncertainty learns until $\nu$=3.1 (last).}
    \label{fig:ka_env_CL}
\end{figure}

In Fig. \ref{fig:barplots_keep_away}, we show bar plots indicating the maximum noise level at which the model converged for different training settings. We define that the model has converged if the success rate of the agent is greater than 90\%. In each of the sub-figures, the parameter mentioned on the Y-axis is progressively increased until failure while the other parameters are set to zero. Check figure \ref{fig:ka_baseline} and \ref{fig:ka_env_CL} for reward graphs.

\textbf{Physical Deception environment:} This is a mixed cooperative and competitive task. There are 2 collaborative agents, 2 landmarks, and 1 adversary. Both the collaborative agents and the adversary want to reach the target, but only collaborative agents know the correct target. The collaborative agents should learn a policy to cover all landmarks so that the adversary does not know which one is the true target.

\begin{figure}[!htbp]
    \centering
    \begin{subfigure}{0.3\linewidth}
        \includegraphics[width=1\linewidth]{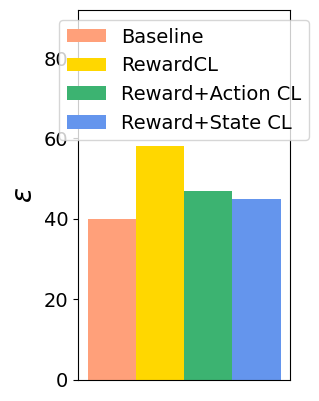}
        \caption{Highest $\epsilon$ for which convergence is achieved for different training strategies.}
        \label{fig:reward_barplot_tag}
    \end{subfigure}
    \hfill
    \centering
    \begin{subfigure}{0.3\linewidth}
        \includegraphics[width=1\linewidth]{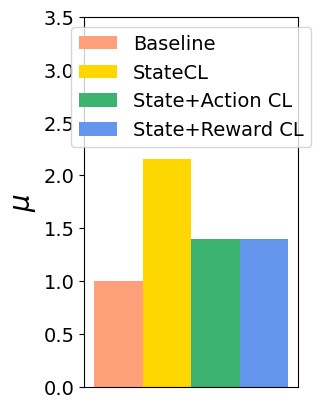}
        \caption{Highest $\mu$ for which convergence is achieved for different training strategies.}
        \label{fig:state_barplot_tag}
    \end{subfigure}
    \hfill
    \centering
    \begin{subfigure}{0.3\linewidth}
        \includegraphics[width=0.95\linewidth]{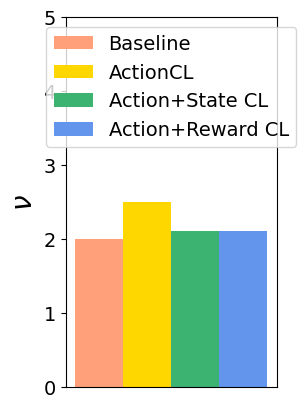}
        \caption{Highest $\nu$ for which convergence is achieved for different training strategies.}
        \label{fig:action_barplot_tag}
    \end{subfigure}
    \caption{\textbf{Physical Deception.} The highest noise values that lead to convergence (agent success rate $>$ 90\%) are shown above. We clearly see that CL applied to a single parameter does better than the baseline training with no CL. When CL is applied to multiple parameters, the highest noise value is usually lower than applying CL to a single parameter. However, the same model now supports uncertainty in two different parameters for a slight reduction in the noise value hence is useful.}
    \label{fig:barplots_pd}
\end{figure}

In Fig. \ref{fig:barplots_pd}, we again show barplots indicating the maximum noise level at which the model converged (success rate $>$ 90\%) for different training settings. In each of the sub-figures, the parameter mentioned on the Y-axis is progressively increased until failure while the other parameters are set to zero.

\vspace{-3mm}
\section{Conclusion and Future Work}
We investigate using curriculum learning to develop a robust multi-agent reinforcement learning model for multi-modal environment uncertainty. For this, we develop an efficient curriculum that slowly increases uncertainty of the uncertain parameters and finally achieves state-of-the-art robustness. It has surpassed baseline for state, reward and action uncertainty significantly both in the case of single uncertainty and multi-modal (two) uncertainty. We show results on three environments to show the validity of our method.

We do not show results for three uncertainty combined because it makes the learning difficult and reduces the final robustness. So as future work we would like to develop a method that can combine all three uncertainties. We would also like to develop conditional theoretical guarantees on the existence of Nash Equilibrium for MARL with multi-modal uncertainty. Finally, we would like to test our robust MARL model for its sim-to-real performance.

\newpage
\bibliography{iclr2024_conference}
\bibliographystyle{iclr2024_conference}
\newpage
\section{Environment Details and Training Parameter Details}
\label{train_details}
\textbf{Cooperative navigation (CN):} This is a cooperative game. There are 3 agents and 3 landmarks. Agents are rewarded based on how far any agent is from each landmark. Agents are penalized if they collide with other agents. So, agents have to learn to cover all the landmarks while avoiding collisions.

\textbf{Keep away (KA):} This is a competitive task. There is 1 agent, 1 adversary, and 1 landmark. The agent knows the position of the target landmark and wants to reach it. The adversary is rewarded if it is close to the landmark and if the agent is far from the landmark. The adversary should learn to push the agent away from the landmark.

\textbf{Physical deception (PD):} This is a mixed cooperative and competitive task. There are 2 collaborative agents, 2 landmarks, and 1 adversary. Both the collaborative agents and the adversary want to reach the target, but only collaborative agents know the correct target. The collaborative agents should learn a policy to cover all landmarks so that the adversary does not know which one is the true target.

\section{Reward Plots for Experiments}
\subsection{Cooperative Navigation Environment}

\begin{figure}[htp]
    \centering
    \begin{subfigure}{0.45\linewidth}
        \includegraphics[width=\linewidth]{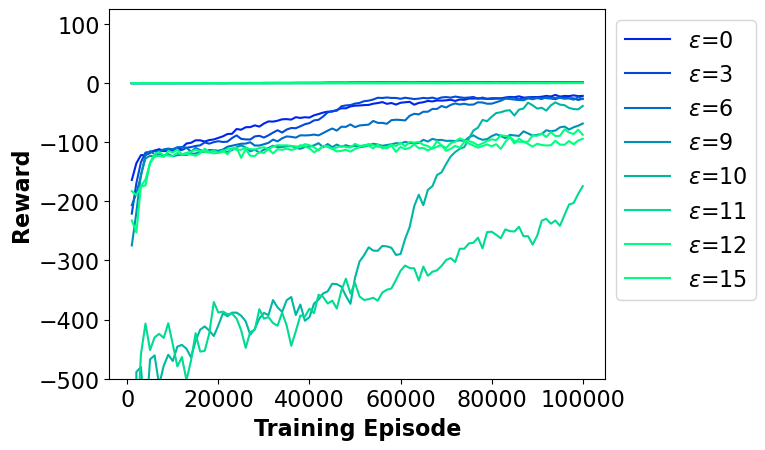}
        \caption{Reward plot for baseline for various $\epsilon$ in reward uncertainty. Baseline was able to learn only till $\epsilon$ = 9.}
    \end{subfigure}
    \hfill
    \begin{subfigure}{0.45\linewidth}
        \includegraphics[width=\linewidth]{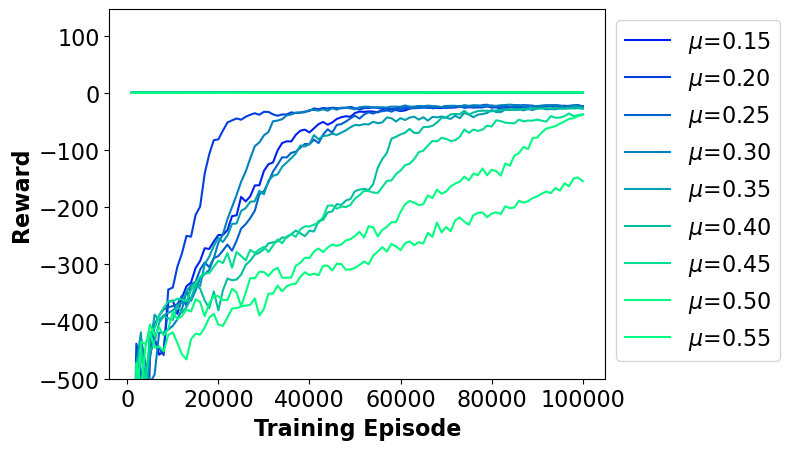}
        \caption{Reward plot for baseline for various $\mu$ in state uncertainty. The baseline was able to learn only till $\mu$ = 0.5.}
    \end{subfigure}
    \hfill
    \begin{subfigure}{0.45\linewidth}
        \includegraphics[width=\linewidth]{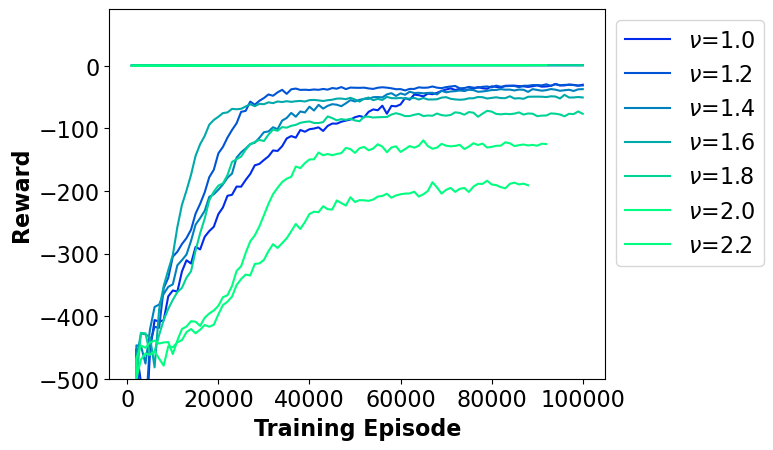}
        \caption{Reward plot for baseline for various $\nu$ in action uncertainty. The baseline was able to learn only till $\mu$ = 2.0.}
    \end{subfigure}
    \caption{\textbf{Cooperative Navigation: Baseline Performance.} Rewards vs training time. Reward uncertainty shows good performance until $\epsilon$=9 (left), state uncertainty shows good performance until $\mu$=0.5 (middle) and action uncertainty shows good performance until $\nu$=2.0 (right).}
\end{figure}

\subsection{Keep Away Environment}

\newpage
\section{Nash Equilibrium for state uncertainty in MARL}
\label{NE_state}
A nice proof for the conditional existence of Nash equilibrium is done in \cite{State2023} for the case of state uncertainty. They define the following robust Markov game,

$$ \mathcal{G} = \{\mathcal{N}, \mathcal{M}, \{\mathcal{S}^i\}_{i \in \mathcal{N}}, \{\mathcal{A}^i\}_{i \in \mathcal{N}}, \{\mathcal{B}^i\}_{i \in \mathcal{N}}, \{r^i\}_{i \in \mathcal{N}}, p, \gamma \}$$

$\mathcal{N} = \{1, 2, ..., N\}$ is the set of $N$ agents and $\mathcal{M} = \{\bar{1}, \bar{2}, ..., \bar{N}\}$ is the corresponding set of $N$ adversaries. $\gamma \in [0,1)$ is the discount factor. $S = S_1 \times S_2 ... \times S_{N}$ is the joint state space. $A = A_1 \times A_2 ... \times A_{N}$ is the joint action space. $p: S \times A \rightarrow \Delta(S)$ are the state transition probabilities. $r^i$ is the reward function for each agent. Every agent $i$ is associated with an adversary $\bar{i}$. The adversary perturbs the true state of each agent $s^i \in S^i$ by producing an action $b^i \in B^i$. The perturbed state $\bar{s}^i = f(s^i, b^i)$ where f is a unique bijection given the state $s^i$.

The Markov game $\mathcal{G}$ is shown to be equivalent to a zero-sum two-person extensive-form game with finite strategies and perfect recall in \cite{State2023}. 

\subsection{Extensive-form game}

An extensive-form game (EFG) is a tree-based representation of a game. An EFG has one root node which indicates the start of the game. Each node branches out into multiple children nodes and each branch represents one possible action. The leaf nodes indicate the end of the game and contain the pay-off/reward for the actions specified by the path from the root node to the leaf node.

The robust optimization equation can be decomposed into a two-player EFG. The first player is the nature/combined adversary who selects the perturbed state and the next player is the combined agent which chooses the best action according to the policy to be learned. The nature player has $|\mathcal{\bar{S}}|$ possible choices for the action and the agent player has $|\mathcal{A}|$ choices where $A = A^{1} \times A^{2} \times ... \times A^{n}$ i.e. the space of all possible actions for all agents. The reward for the nature player is the negative of the reward obtained by the action taken by the combined agent.  \\

The Bellman equation for the above game $\mathcal{G}$ is written as below:

$$ v^i(s) = \max_{\pi^i} \min_{\rho^i} \mathbb{E} \left[ \sum_{s^{'} \in S} p(s^{'}|s, a, b)[r^{i}(s, a, b)+\gamma v^{i}(s)] | a \sim \pi(\cdot|\bar{s}),b \sim \rho(\cdot|s)\right] $$

In order for the NE (and the optimal solution to the above equation) to exist, below conditions need to be met:
\begin{itemize}
    \item $\mathcal{S}^i$, $\mathcal{A}^i$ and $\mathcal{B}^i$ must be finite sets $\forall i \in \mathcal{N}$.
    \item $|r^{i}(s, a, b)| < M_{i} < M < \infty \; \forall \; i \in N, a \in A, b \in B$ and $s \in S$
    \item Stationary reward and transition probabilities
    \item f is a bijection for a given $s^{i}$
    \item All agents have the same reward function.
\end{itemize}

\section{NE for reward and transition dynamics uncertainty in MARL}
\label{NE_reward}

In this section, we show how uncertainty in reward and transition dynamics is handled in a multi-agent setting. We follow \cite{marl_uncern} and use the following definition of robust Markov game.

$$\mathcal{\bar{G}} =  \langle \mathcal{N}, \mathcal{S}, \{ \mathcal{A}^{i} \}_{i \in \mathcal{N}}, \{ \mathcal{\bar{R}}^{i}_{s} \}_{(i,s) \in \mathcal{N} \times \mathcal{S}}, \{ \mathcal{\bar{P}}_{s} \}_{s \in \mathcal{S}},  \gamma \rangle$$ 

Note: In this proof following \cite{marl_uncern} $s_t$ denotes the system state and not the individual agent state.
The expected return in case of multi-agent RL with no uncertainty for $i^{th}$ agent is -

$$V^{i}_{\pi}(s) = \mathbb{E} [ {\sum_{t=0}^{\infty}} \gamma^{t} r^{i}_{t} | s_{0} = s, a^{i}_{t} \sim \pi^{i}(.|s_{t}), a^{-i}_{t} \sim \pi^{-i}(.|s_{t}) ]$$

where $-i$ represents the indices of all agents except agent $i$, and $\pi^{-i} = \Pi_{i \neq j} \pi_{j}$ refers to the joint policy of all agents except agent $i$. In order to find the optimal robust value function for the single agent the other agent policies are considered stationary. Since all policies are evolving continuously and expected return is dependent on all agent policies, one commonly used solution for optimal policy $\pi^{*} = \{\pi_{1}^{*}, \pi_{2}^{*}, \dots \pi_{N}^{*} \}$ is Nash equilibrium. Non-stationarity is also one of the main reasons for difficulty in MARL convergence as compared to single agent RL which also reflects when uncertainty is added.

We now introduce uncertainty in rewards and transition dynamics. Thus, the desired policy should now not only be able to play against other agents’ policies but also robust to the possible uncertainty of the MARL model. Each player considers a distribution-free Markov game to be played using robust optimization. To find the optimal value function we focus on the following idea from \cite{marl_uncern}. If the player knows how to play in the robust Markov game optimally starting from the next stage on, then it would play to maximize not only the worst-case (minimal) expected immediate reward, due to the model uncertainty set at the current stage, but also the worst-case expected reward incurred in the future stages. Formally, such a recursion property leads to the following Bellman-type equation:

$$\bar{V}^{i}_{*}(s) = \max_{\pi^{i}(.|s)} \min_{ \begin{array}{c}\scriptstyle \bar{P}(.|s,.) \in \mathcal{\bar{P}}_{s} \\[-4pt] \scriptstyle \bar{R}^{i}_{s} \in \mathcal{\bar{R}}^{i}_{s} \end{array}} \sum_{a \in A} \prod_{j=1}^{N} \pi^{j}(a^{j} |s)(\bar{R}^{i}(s,a) + \gamma \sum_{s' \in S} \bar{P}(s'|s,a) \bar{V}^{i}_{*}(s'))$$

The corresponding joint policy $\pi^{*} = \{\pi^{1}, \pi^{2} \dots \pi^{N}\}$ is robust Markov perfect Nash equilibrium. 


\section{Proof for Theorem 1}
\label{theorem1}

Lets define the non-linear operator on $\mathcal{L}$ such that, 


$$\mathcal{L}^{i} v^{i} (s) = \max_{\pi^{i}(.|s^{i})} \min_{\rho} \left[ \sum_{a \in \mathcal{A}}  \bar{R}^{i}(s, \bar{a}) + \gamma \sum_{s'\in S} \bar{P}(s'|s, \bar{a}) v^{i}(s') \right] $$, where $\rho = \{ \bar{P}, \bar{R}, \bar{s}, \bar{a} \}$

We can think of $\rho$ as adversarial strategy that is playing against the good policy $\pi$ by selecting the values $\{ \bar{P}, \bar{R}, \bar{s}, \bar{a} \}$ from their respective uncertainty sets such that it minimises the expected return.

Let $u$ and $v$ be two value functions in $\mathbb{V}$. Let $\{ \pi_{*}^{u}, \rho_{*}^{u} \}$ and $\{ \pi_{*}^{v}, \rho_{*}^{v} \}$ be two different Nash Equilibrium with respect to  $\mathcal{\bar{G}}_{general}$.

$$\mathcal{L}^{i} v^{i} (s) =  \sum_{a \in \mathcal{A}}  \bar{R}^{i}(s, \bar{a})_{(\pi_{*}^{v}, \rho_{*}^{v})} + \gamma \sum_{s'\in S} \bar{P}(s'|s, \bar{a})_{(\pi_{*}^{v}, \rho_{*}^{v})} v^{i}(s') $$, where $\rho_{*}^{v} = \{ \bar{P}, \bar{R}, \bar{s}, \bar{a} \}$ is the optimal value that minimises the value function equation. 
$$\mathcal{L}^{i} u^{i} (s) =  \sum_{a \in \mathcal{A}}  \bar{R}^{i}(s, \bar{a})_{(\pi_{*}^{u}, \rho_{*}^{u})} + \gamma \sum_{s'\in S} \bar{P}(s'|s, \bar{a})_{(\pi_{*}^{u}, \rho_{*}^{u})} u^{i}(s') $$, where $\rho_{*}^{u} = \{ \bar{P}, \bar{R}, \bar{s}, \bar{a} \}$ is the optimal value that minimises the value function equation. 

Its intuitive that optimal $\pi_{*}$ maximizes the above equation, whereas optimal $\rho_{*}$ minimises the above equation. Therefore we can write the following equation,

$$ \sum_{a \in \mathcal{A}}  \bar{R}^{i}(s, \bar{a})_{(\pi_{*}^{u}, \rho_{*}^{v})} + \gamma \sum_{s'\in S} \bar{P}(s'|s, \bar{a})_{(\pi_{*}^{u}, \rho_{*}^{v})} v^{i}(s') \leq \mathcal{L}^{i} v^{i} (s) \leq  \sum_{a \in \mathcal{A}}  \bar{R}^{i}(s, \bar{a})_{(\pi_{*}^{v}, \rho_{*}^{u})} + \gamma \sum_{s'\in S} \bar{P}(s'|s, \bar{a})_{(\pi_{*}^{v}, \rho_{*}^{u})} v^{i}(s') $$

$$ \sum_{a \in \mathcal{A}}  \bar{R}^{i}(s, \bar{a})_{(\pi_{*}^{v}, \rho_{*}^{u})} + \gamma \sum_{s'\in S} \bar{P}(s'|s, \bar{a})_{(\pi_{*}^{v}, \rho_{*}^{u})} u^{i}(s') \leq \mathcal{L}^{i} u^{i} (s) \leq  \sum_{a \in \mathcal{A}}  \bar{R}^{i}(s, \bar{a})_{(\pi_{*}^{u}, \rho_{*}^{v})} + \gamma \sum_{s'\in S} \bar{P}(s'|s, \bar{a})_{(\pi_{*}^{u}, \rho_{*}^{v})} u^{i}(s') $$

(a) Now lets assume, $\mathcal{L}^{i} v^{i} (s) \leq \mathcal{L}^{i} u^{i} (s) $

\begin{align*}
    0 &\leq \mathcal{L}^{i} u^{i} (s) - \mathcal{L}^{i} v^{i} (s) \\
     &\leq \left[ \sum_{a \in \mathcal{A}}  \bar{R}^{i}(s, \bar{a})_{(\pi_{*}^{u}, \rho_{*}^{v})} + \gamma \sum_{s'\in S} \bar{P}(s'|s, \bar{a})_{(\pi_{*}^{u}, \rho_{*}^{v})} u^{i}(s') \right] - \left[ \sum_{a \in \mathcal{A}}  \bar{R}^{i}(s, \bar{a})_{(\pi_{*}^{u}, \rho_{*}^{v})} + \gamma \sum_{s'\in S} \bar{P}(s'|s, \bar{a})_{(\pi_{*}^{u}, \rho_{*}^{v})} v^{i}(s') \right] \\
     & \leq \gamma \sum_{s'\in S} \bar{P}(s'|s, \bar{a})_{(\pi_{*}^{u}, \rho_{*}^{v})} ( u^{i}(s')  -  v^{i}(s') )  \\
     & \leq \gamma || u^{i}(s')  -  v^{i}(s') ||
\end{align*}

(b) Assuming , $\mathcal{L}^{i} u^{i} (s) \leq \mathcal{L}^{i} v^{i} (s) $ and following the same argument as before we get, 

$$  \mathcal{L}^{i} v^{i} (s) - \mathcal{L}^{i} u^{i} (s) \leq \gamma|| v^{i}(s')  -  u^{i}(s') || $$

Thus, combining (a) and (b), we get,
$$  || \mathcal{L}^{i} v^{i} (s) - \mathcal{L}^{i} u^{i} (s) || \leq \gamma|| v^{i}(s')  -  u^{i}(s') || $$

Thus, $\mathcal{L}^{i}$ is a contraction mapping on $V$

Now since $|| v || =  \sup_{i} || v^{i} || $, we can write the following - 
$$|| \mathcal{L} v  - \mathcal{L} u || = \sup_{i}|| \mathcal{L}^{i} v^{i}  - \mathcal{L}^{i} u^{i}  || \leq \gamma \sup_{i} || v^{i} -  u^{i} || = \gamma|| v - u|| $$

Thus, $\mathcal{L}$ is a contraction mapping on $\mathbb{V}$


\end{document}